\documentclass[lettersize,journal]{IEEEtran}
\usepackage{amsmath,amsfonts}
\usepackage[ruled,vlined]{algorithm2e}
\usepackage{array}
\usepackage[caption=false,font=normalsize,labelfont=sf,textfont=sf]{subfig}
\usepackage{textcomp}
\usepackage{stfloats}
\usepackage{url}
\usepackage{verbatim}
\usepackage{graphicx}
\usepackage{cite}
\usepackage{booktabs}
\usepackage{multirow} 
\usepackage{makecell}
\usepackage{pifont}
\usepackage{natbib}  
\setcitestyle{numbers,sort&compress}
\usepackage{caption} 
\usepackage[caption=false, font=normalsize, labelfont=sf, textfont=sf]{subfig}
\begin{document}

\title{Vision SmolMamba: Spike-Guided Token Pruning for Energy-Efficient Spiking State-Space Vision Models}
\author{Dewei Bai, Hongxiang Peng, Yunyun Zeng, Ziyu Zhang, Hong Qu, Yi Zhang

\thanks{This work has been submitted to the IEEE for possible publication. Copyright may be transferred without notice, after which this version may no longer be accessible.}
\thanks{Dewei Bai, Hongxiang Peng, Yunyun Zeng, Ziyu Zhang, and Hong Qu are with the School
of Computer Science and Engineering, University of Electronic Science and Technology of China, Chengdu 610054, China.}
\thanks{Zhang Yi is with the College of Computer Science, Sichuan University, Chengdu 610065, China.} }



\maketitle

\begin{abstract}
Spiking Transformers have shown strong potential for long-range visual modeling through spike-driven self-attention.  However, their quadratic token interactions remain fundamentally misaligned with the sparse and event-driven nature of spiking neural computation.
To address this limitation, we propose Vision SmolMamba, an energy-efficient spiking state-space architecture that integrates spike-driven dynamics with linear-time selective recurrence. The key idea is a Spike-Guided Spatio-Temporal Token Pruner (SST-TP), which estimates token importance using both spike activation strength and first-spike latency.  This mechanism progressively removes redundant tokens while preserving salient spatio-temporal information, enabling efficient scaling with token sparsity.
Based on this mechanism, the proposed SmolMamba block incorporates spike events directly into bidirectional state-space recurrence, forming a spiking state-space vision backbone for efficient long-range modeling. Extensive experiments on both static and event-based benchmarks, including ImageNet-1K, CIFAR10/100, CIFAR10-DVS, and DVS128 Gesture, demonstrate that Vision SmolMamba consistently achieves superior accuracy–efficiency trade-offs.  In particular, it reduces the estimated energy cost by at least 1.5× compared with prior spiking Transformer baselines and a Spiking Mamba variant while maintaining competitive or improved accuracy.
These results demonstrate that combining spike-guided token sparsity with state-space modeling offers a scalable and energy-efficient paradigm for spiking vision systems.
\end{abstract}


\section{Introduction}
\begin{figure}[t]
    \centering
    \includegraphics[width=1.0\linewidth]{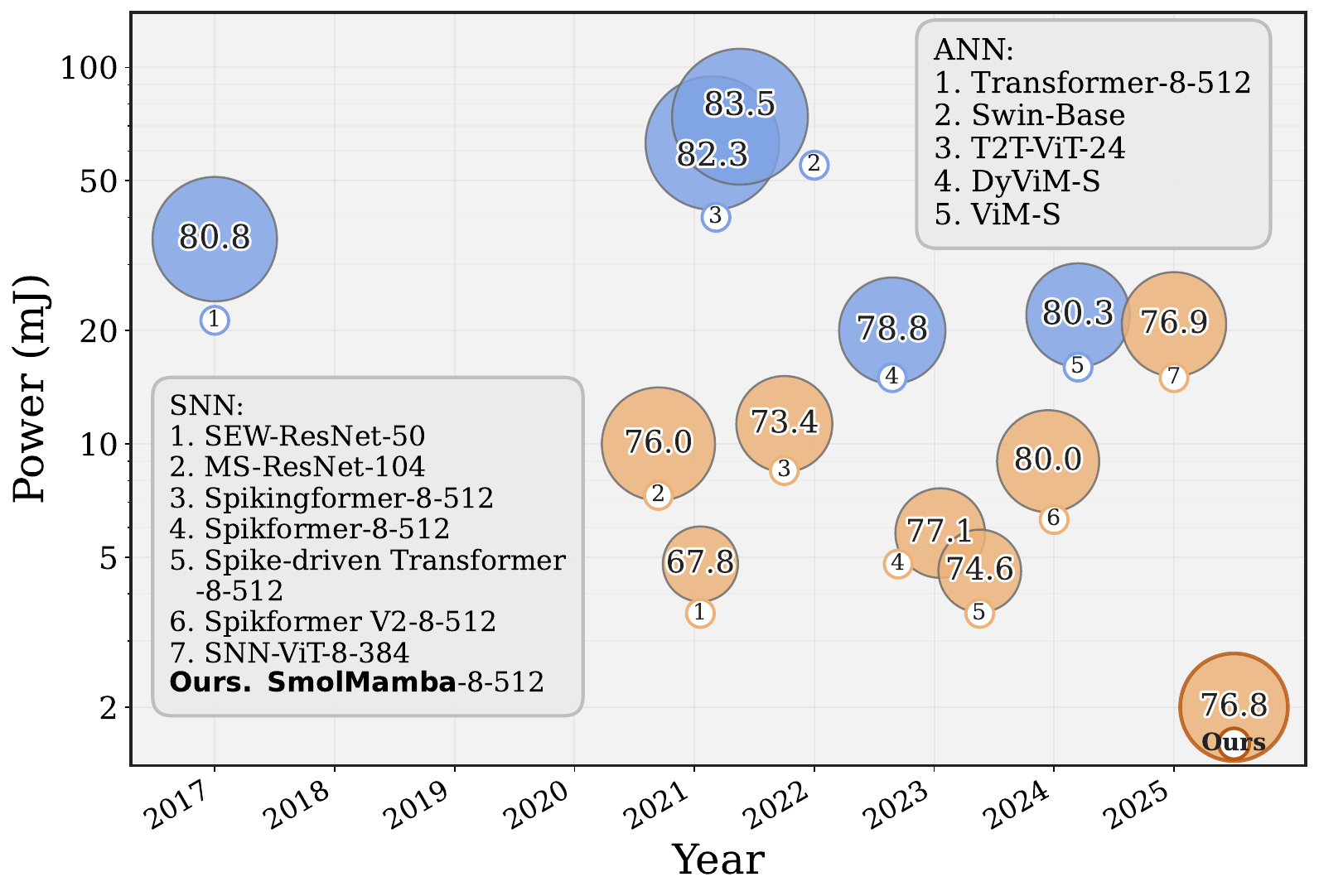} 
    \caption{
        The estimated energy–accuracy landscape of ANN and SNN models on ImageNet-1K. While modern ANN models (e.g., Swin-Base, T2T-ViT-24) achieve high accuracy, they generally lie in a high estimated-energy regime. In contrast, contemporary SNN models (e.g., Spikingformer, Spikformer V2) demonstrate competitive accuracy with substantially lower estimated energy cost, highlighting SNNs as a promising pathway for energy-efficient inference.
    }
    \label{fig:energy_accuracy_tradeoff}
\end{figure}

\IEEEPARstart{S}{piking} Neural Networks (SNNs), inspired by biological neural systems, process information through discrete spike events rather than continuous activations. This event-driven mechanism enables sparse and asynchronous computation, leading to remarkable energy efficiency and strong temporal modeling capability on neuromorphic hardware~\cite{nunesSpikingNeuralNetworks2022,bouvier2019spiking}. Despite these advantages, extending SNNs to large-scale visual tasks remains challenging. Their sparse firing patterns and local receptive-field operations often limit representational capacity, particularly in capturing long-range spatial dependencies essential for complex visual understanding. As illustrated in Fig.~\ref{fig:energy_accuracy_tradeoff}, the energy–accuracy trend in recent years shows that while ANN architectures (e.g., Swin-Base and T2T-ViT) dominate in accuracy, SNN counterparts can achieve competitive performance with substantially lower energy consumption, highlighting their potential for energy-efficient intelligence.

Recent studies have explored two primary paradigms for visual SNNs: convolutional SNNs and Spiking Vision Transformers (Spiking ViTs). Convolutional SNNs~\cite{fang2021incorporating} efficiently capture local spatial features but struggle with global modeling due to their inherently local operations. In contrast, Spiking Transformers~\cite{yao2023spike,yu2025spikingvit} introduce spike-driven self-attention to capture global spatio-temporal dependencies, significantly improving representational power. However, their quadratic attention complexity and dense matrix operations conflict with the sparse and event-driven nature of spiking computation, limiting both scalability and energy efficiency.

Recently, the Mamba architecture~\cite{gu2024mamba}, derived from structured state-space models (SSMs)~\cite{fu2023hungry}, has emerged as an efficient alternative for long-sequence modeling. By replacing quadratic self-attention with linear recurrence, Mamba achieves efficient long-range dependency modeling with linear-time complexity, making it a promising candidate for energy-efficient neuromorphic adaptation.

The integration of SSMs with SNNs has recently attracted increasing attention for efficient sequence modeling. Foundational studies~\cite{shen2025spikingssms,bal2024p} have begun to bridge their fundamental differences by combining the long-range dependency modeling capability of SSMs with the event-driven sparsity of SNNs. Extensions to specialized domains such as 3D perception, temporal video grounding, and event-based vision~\cite{wu2025efficient,li2024spikemba,chen2024spikmamba,zubic2024state} further demonstrate the potential of this synergy. Nevertheless, many existing approaches still rely on relatively dense gating or attention mechanisms, which partially undermine spike-driven sparsity and limit computational efficiency. Consequently, developing a unified and energy-efficient framework that seamlessly integrates spike-based computation with state-space long-sequence modeling for general frame-based vision remains an open challenge.

To address this challenge, we propose Vision SmolMamba, an energy-efficient spiking state-space architecture that performs progressive spatio-temporal token evolution within a linear-time recurrence framework. The main contributions of this work are summarized as follows:

\begin{itemize}

\item We introduce the first spiking state-space architecturefor vision tasks, bridging event-driven spiking computationwith linear-time state-space modeling.

\item We propose a spike-guided spatio-temporal token pruning mechanism that exploits both spike activity and first-spike latency to estimate token importance.

\item We demonstrate that the proposed architecture achieves a significantly improved accuracy–energy trade-off on both static and event-based vision benchmarks .

\end{itemize}

To our knowledge, Vision SmolMamba is the first framework that combines spike-driven dynamics with state-space recurrence through progressive token pruning, providing an energy-efficient framework for scalable long-range visual modeling in spiking neural networks.

\section{Related Work}
\subsection{Spiking Neural Networks for Vision Tasks}

Spiking Neural Networks (SNNs) have been applied across diverse computer vision tasks due to their event-driven efficiency. Their scope has expanded from 2D images to 3D point clouds \cite{ren2023spiking} and neuromorphic sensor data, where spike cues enhance tasks such as motion deblurring \cite{chen2023enhancing}. A major line of work scales SNNs through deep convolutional backbones, evolving from ANN-to-SNN conversion \cite{hu2021spiking} to direct-training approaches like SEW ResNet \cite{fang2021deep}, further boosted with attention mechanisms \cite{yao2023attention}.
While convolutional SNNs capture local features effectively, they struggle with long-range dependencies. Transformer-based SNNs address this via spike-form self-attention, replacing expensive Softmax attention with spike-friendly operations such as Q/K/V spikes \cite{zhou2023spikformer} or mask-based additions \cite{zheng2023capture}. Architectural refinements such as spiking residuals \cite{zhou2023spikingformer} improve hardware-friendliness. Wang et al.~\cite{wangspiking} further introduced a Saccadic Spike Self-Attention (SSSA) mechanism inspired by biological saccadic attention, which enhances spatial relevance estimation and temporal interactions in spiking vision transformers.  SparseSpikFormer~\cite{liu2024sparsespikformer} introduces a lightweight token selector that prunes background tokens based on spike firing rates, enabling token-level sparsity and reducing redundant token interactions within spiking transformers. Despite these improvements, attention-based SNN architectures still rely on quadratic token interactions, which limits their scalability when processing high-resolution visual inputs. This limitation motivates the exploration of alternative architectures capable of modeling long-range dependencies with lower computational complexity.

\subsection{Vision Mambas}

Modern Vision Mambas evolved from State Space Models (SSMs), developed to overcome the $O(N^2)$ complexity of Transformers in long-sequence modeling. Initial models like S4 \cite{gu2022efficiently} and S5 \cite{smith2023simplified} demonstrated near-linear complexity on long-range benchmarks but struggled with content-dense modalities. The breakthrough came with Mamba \cite{gu2024mamba}, which introduced an input-dependent selection mechanism. This allowed Mamba to selectively manage information, achieving linear-time scaling and SOTA performance, establishing SSMs as a viable alternative to Transformers.

The core advantages of 1D SSMs, such as linear complexity and parallel-friendly computation, prompted their adaptation as generic backbones for 2D vision. Foundational works like Vision Mamba (Vim) \cite{zhuVisionMambaEfficient2024} used bidirectional Mamba blocks to rival ViT performance efficiently. Concurrently, VMamba \cite{liu2024vmamba} addressed the 1D-to-2D "direction-sensitive issue" by introducing a Cross-Scan Module (CSM).

Subsequent research rapidly refined these architectures. For instance, LocalMamba \cite{65f3abd913fb2c6cf6612b32} introduced windowed scanning to better capture 2D local dependencies. MSVMamba \cite{shi2024multi} tackled multi-scan redundancy with a multi-scale approach, while Mamba-Reg \cite{wang2025mamba} introduced register tokens to mitigate feature map artifacts. And most recently, Dynamic Vision Mamba (DyVM) \cite{wu2025dynamic} analyzed both token- and block-redundancy in Mamba-based vision models and proposed adaptive pruning of tokens plus dynamic block selection, reducing roughly 35.2\% FLOPs for minor accuracy loss. In parallel, analytical works like \cite{han2024demystify} sought to ``demystify" Mamba, identifying its ``forget gate" and ``block design" as the core components of its success by comparing it to linear attention.

Although Vision Mambas achieve competitive results with improved computational efficiency, they are primarily designed for dense artificial neural networks and do not explicitly consider the sparse event-driven characteristics of spiking neural computation.

\subsection{Spiking Neural Networks for SSMs}

The integration of SSMs with SNNs is an emergent field aiming to combine the long-sequence modeling of SSMs with the sparsity of SNNs. Foundational work addresses their core computational conflicts. \cite{shen2025spikingssms} proposed SpikingSSMs to reconcile SSM parallel scans with SNN sequential dynamics, enabling parallel training. Concurrently, \cite{bal2024p} introduced a probabilistic SSM-based framework using stochastic sampling to effectively model long-range dependencies on benchmarks like the LRA.

Applications of this synergy have emerged in various domains. These include Spiking Point Mamba (SPM) for 3D point cloud processing \cite{wu2025efficient} and SpikeMba for multi-modal Temporal Video Grounding (TVG) \cite{li2024spikemba}. In vision, explorations are largely confined to asynchronous event-based data. For instance, \cite{zubic2024state} utilized SSMs to robustly handle varying event stream frequencies, while \cite{chen2024spikmamba} introduced SpikMamba, an SNN-Mamba framework for event-based Human Action Recognition (HAR). Currently, some researchers have begun to explore the study of standard (non-pulsed) Mamba models in terms of token redundancy, including token reduction strategies and parameter pruning \cite{zhan2024exploring, zhan2024rethinking, shihab2025efficient}. 

However, these existing optimization techniques are designed for dense, non-spiking architectures and are ill-suited for the sparse, event-driven dynamics unique to SNNs. A critical gap remains in developing efficiency-aware methods specifically tailored for hybrid SNN-SSM architectures. 
To address this limitation, we introduce SmolMamba, an energy-efficient spiking state-space model. We elaborate on this method in the following section.

\section{Preliminary}
\subsection{Leaky Integrate-and-Fire (LIF) Neuron}
We employ the standard Leaky Integrate-and-Fire (LIF) neuron as the spiking nonlinearity.
Let $v(t)$ denote the membrane potential and $I(t)$ the synaptic input current.
The subthreshold dynamics are governed by a first-order low-pass filter:
\begin{equation}
\tau \frac{dv(t)}{dt} = -\,v(t) + I(t),
\label{eq:lif_cont}
\end{equation}
where $\tau>0$ is the membrane time constant. When $v(t)$ reaches the threshold
$V_{\mathrm{th}}$, the neuron emits a spike and the membrane is reset to $V_{\mathrm{reset}}$.

For discrete-time simulation with step $\Delta t$ (we define $\alpha=\Delta t/\tau$),
the membrane update follows an Euler integration scheme:
\begin{equation}
\tilde{v}[t] = \alpha\,v[t-1] + x[t],
\label{eq:lif_tmp}
\end{equation}
\begin{equation}
s[t] = \Theta\!\big(\tilde{v}[t] - V_{\mathrm{th}}\big), \qquad
\Theta(u)=
\begin{cases}
1,& u\ge 0,\\
0,& u<0,
\end{cases}
\label{eq:lif_spk}
\end{equation}
\begin{equation}
v[t] = \tilde{v}[t]\,(1-s[t]) + V_{\mathrm{reset}}\,s[t],
\label{eq:lif_reset}
\end{equation}
where $x[t]$ is the input drive at time step $t$, 
$s[t]\!\in\!\{0,1\}$ is the spike output, and $\tilde{v}[t]$ is the pre-spike potential.
Eqs.~\eqref{eq:lif_tmp}--\eqref{eq:lif_reset} jointly realize the leaky integration, 
threshold firing, and reset process of the LIF neuron.

\paragraph{Notes for training}
During backpropagation, we employ a Sigmoid surrogate gradient to approximate the non-differentiable Heaviside step function $\Theta(\cdot)$ in spiking neurons, while keeping the forward spike generation identical to Eq.~\eqref{eq:lif_spk}.
Specifically, the surrogate function is defined as
\begin{equation}
    \sigma(u) = \frac{1}{1+\exp(-\alpha u)},
\end{equation}
where $\alpha$ controls the sharpness of the gradient; we set $\alpha=4$ in all experiments.
All LIF neuron hyperparameters $(\tau=2.0, V_{\mathrm{th}}=0.5, V_{\mathrm{reset}}=0.0)$ are kept fixed unless otherwise specified.

\subsection{Selective SSM}
In SmolMamba Block, $\operatorname{SSM}(\cdot)$ denotes a selective, input-dependent SSM as in Mamba~\cite{gu2024mamba}.
Given a per-step feature $u_k$ produced by the spiking frontend, we form input-dependent parameters
\begin{equation}
\Delta_k=\phi_\Delta(u_k),\quad B_k=\phi_B(u_k),\quad C_k=\phi_C(u_k),
\end{equation}
and adopt a diagonal state matrix $A=\mathrm{diag}(\lambda)$ for stable, elementwise updates (\(\mathrm{Re}(\lambda)<0\)).
Under ZOH discretization, the per-step coefficients become
\begin{equation}
\overline{A}_k=\exp(\Delta_k \odot \lambda),\qquad
\overline{B}_k=\frac{\exp(\Delta_k \odot \lambda)-\mathbf{1}}{\lambda}\odot B_k,
\end{equation}
leading to the selective recurrence
\begin{equation}
h_k=\overline{A}_k \odot h_{k-1}+\overline{B}_k \odot x_k,\qquad
y_k=C_k \odot h_k + D x_k,
\end{equation}
with $\odot$ the elementwise product and safe division understood channel-wise (use \(\lambda\) bounded away from \(0\) or add \(\varepsilon\)).
We use a bidirectional variant by scanning both the original and the reversed sequence and summing the outputs (Eq.~\ref{eq:SSM}).
The selective scan runs in linear time and memory w.r.t. sequence length \(N\), i.e., \(\mathcal{O}(N C)\).

\section{Method}
\begin{figure*}[t]
  \centering
    \subfloat[]{
        \includegraphics[width=0.6\linewidth]{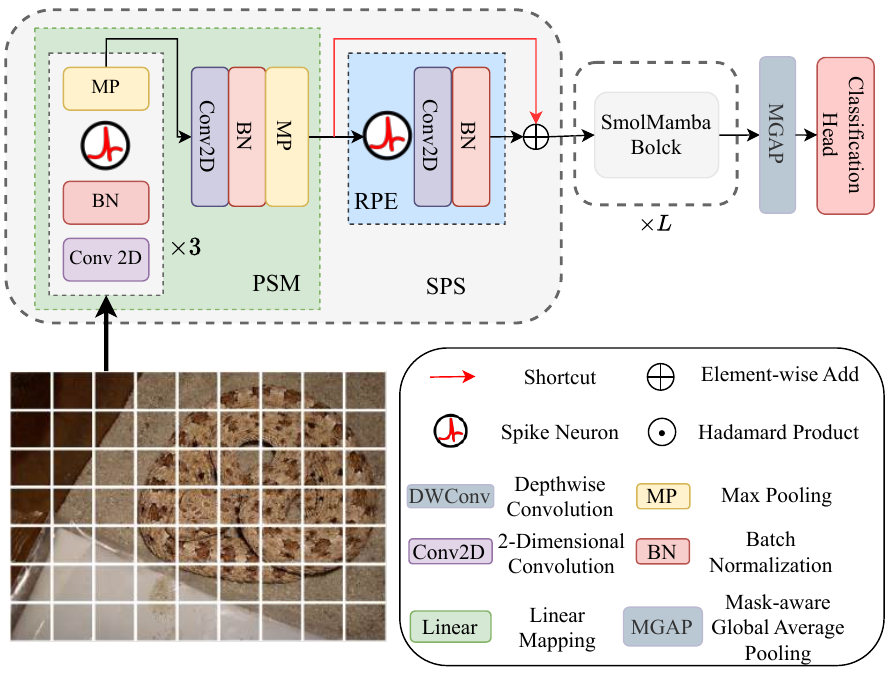}
        \label{fig:overall}
    }
  \hfill
  \subfloat[]{
        \includegraphics[width=0.325\linewidth]{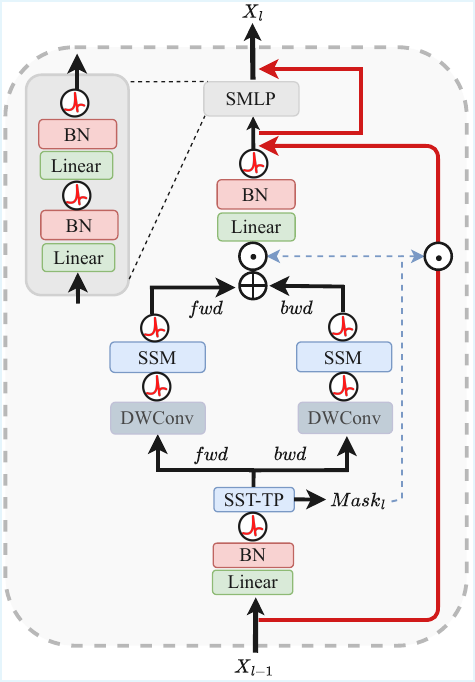}
        \label{fig:smolmamba}
    }
  \caption{Overview of Vision SmolMamba. 
(a) The overall architecture: spike-form visual patches generated with SPS are processed by a stack of SmolMamba blocks that progressively prune redundant tokens and perform spiking state-space modeling. A Mask-aware Global Average Pooling (MGAP) and a linear Classification Head (CH) aggregate the final spiking representation. 
(b) SmolMamba block: input $X_{l-1}$ is processed by an initial projection (BN, Linear, Spike Neuron). The SST-TP module generates the mask $\text{Mask}_l$. This mask (blue dotted line) is applied in two key locations. One is to modulate the fused output of the fwd/bwd SSM branches. The other is to modulate the residual shortcut $X_{l-1}$.}

  \label{fig:all}
\end{figure*}

\subsection{Overall Architecture}
An overview of the proposed architecture \textsc{Spiking Molting Mamba} (Vision SmolMamba) is shown in Figure~\ref{fig:overall}.
For an $M$-class visual classification task, we consider an input 2D image sequence of length $T$, denoted as $I \in \mathbb{R}^{B \times T \times C \times H \times W}$, where $B$, $T$, $C$, $H$, and $W$ denote the batch size, timestep, channel, height, and width dimensions, respectively.
We adopt the \textsc{Spiking Patch Splitting (SPS)} module from~\cite{yao2023spike} to project and divide the input into a sequence of $N_0$ flattened spike-form patches $x \in \mathbb{B}^{B \times T  \times C \times N_0}$, each with a feature dimensionality of $D$.

A spike-form \textsc{relative position embedding (RPE)} is generated by a convolutional layer 
followed by batch normalization and a spiking nonlinearity, and is then added to $x$ to obtain $X_0$. 
The resulting feature sequence $X_0$ is subsequently passed through the 
$L$-block \textsc{SmolMamba block}, which serves as the core component of our architecture. 
Each SmolMamba block ---illustrated in Fig.~\ref{fig:smolmamba}---employs a serial residual structure consisting of a dynamic pruner, spiking SSM module, spiking MLP and Mask BN.

During the classification stage, a unified \textsc{mask-aware Global Average Pooling (MGAP)}
is applied to the encoder output, and the aggregated representation is then fed into 
a linear \textsc{Classification Head (CH)} for the final prediction.

The overall architecture can be expressed as:

\begin{equation}
\begin{aligned}
x = \operatorname{SPS}(I) &\in \mathbb{B}^{B \times T \times C \times N_0},\\
\mathrm{RPE} = \mathcal{SN}\!\left(\mathrm{BN}\!\left(\operatorname{Conv2d}(x)\right)\!\right)
    &\in \mathbb{B}^{B \times T \times C \times N_0},\\
X_{0} = x + \mathrm{RPE} &\in \mathbb{B}^{B \times T \times C \times N_0},\\
X_{l}, \text{Mask}_{l} = \operatorname{SmolMamba}_{l}(X_{l-1}) 
    &\in \mathbb{B}^{B \times T \times C \times N_l},\\
Y = \operatorname{CH}\!\left(\operatorname{MGAP}\!\left(X_{L}, \operatorname{Mask}_{L}\right)\!\right)
    &\in \mathbb{B}^{B \times M}.
\end{aligned}
\end{equation}

These equations collectively describe the forward propagation process of the proposed Vision SmolMamba framework, from spike-form patch extraction to the final classification output.

\begin{figure}[t]
    \centering
    \includegraphics[width=1.025\linewidth]{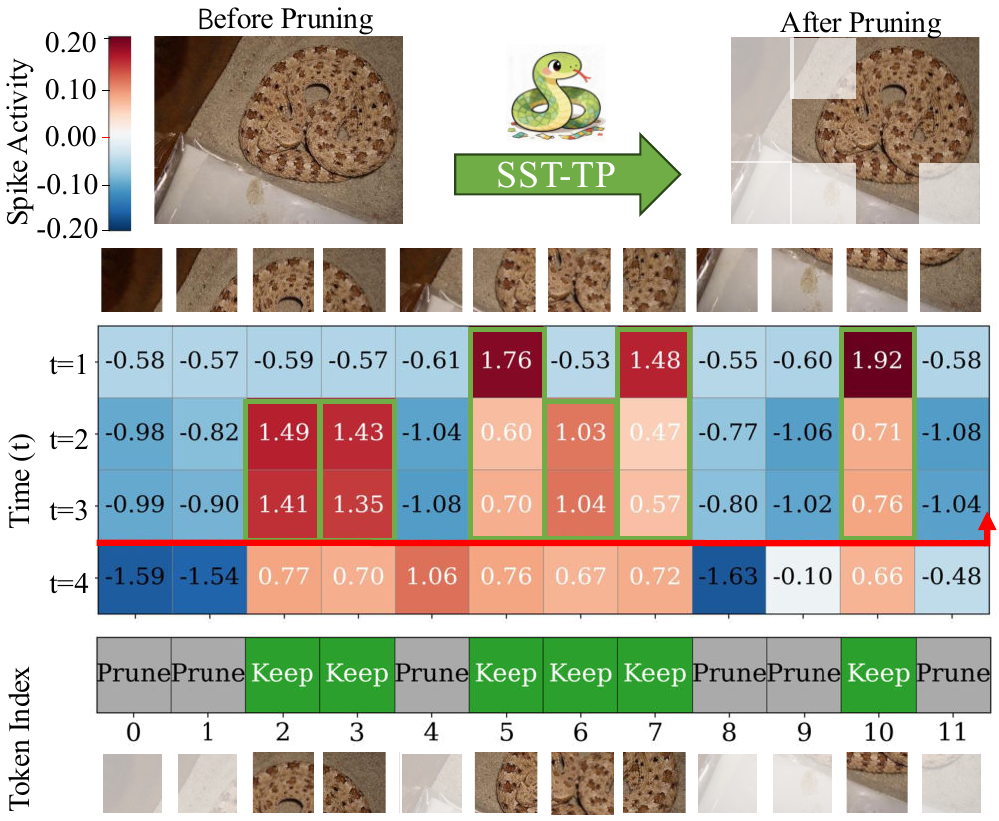}
    \caption{Illustration of the proposed Spike-Guided Spatio-Temporal Token Pruner (SST-TP). 
    The pruner jointly evaluates each token’s spatial spike activity and temporal salience to dynamically remove redundant ones. 
    A token is pruned only when it simultaneously exhibits weak spike activity (below the spatial threshold $\theta_l = 0$) and delayed first-spike latency (beyond the temporal threshold $\phi_l = 3$). 
    This conjunctive criterion ensures that tokens containing either strong or timely spikes are preserved.}
    \label{fig:dsttp}
\end{figure}

\begin{algorithm}[t]
\caption{Spike-Guided Spatio-Temporal Token Pruner (SST-TP)}
\label{alg:dst_tp}
\KwIn{$O_l\!\in\!\mathbb{B}^{B\times T\times C\times N_{l-1}}$, latency threshold $\phi_l$}
\KwOut{$P_l\!\in\!\mathbb{B}^{B\times T\times C\times N_l}$,\ \ $\mathrm{Mask}_l\!\in\!\mathbb{B}^{B\times N_l}$}

\textbf{(1) Channel-wise spike activity \& Z-score normalization}\\
$\bar{A}_l \gets \frac{1}{C}\sum_{c=1}^{C} O_l[:,:,c,:]\ \in\ \mathbb{R}^{B\times T\times N_{l-1}}$\\
$\mu_l[b,t] \gets \operatorname{Mean}_{n}\big(\bar{A}_l[b,t,n]\big)\ \in\ \mathbb{R}^{B\times T}$\\
$\sigma_l[b,t] \gets \operatorname{Std}_{n}\big(\bar{A}_l[b,t,n]\big)\ \in\ \mathbb{R}^{B\times T}$\\
$\tilde{A}_l[b,t,n] \gets \dfrac{\bar{A}_l[b,t,n]-\mu_l[b,t]}{\sigma_l[b,t]+\varepsilon}\ \in\ \mathbb{R}^{B\times T\times N_{l-1}}$

\textbf{(2) Per-timestep Z-score based spatial indicator}\\
$M_l^{\text{spatial}} \gets \mathbb{I}(\tilde{A}_l>0)\ \in\ \mathbb{B}^{B\times T\times N_{l-1}}$

\textbf{(3) First-spike latency based spatio-temporal mask}\\
$t^\star(b,n) \gets \min\{\, t \mid M_l^{\text{spatial}}(b,t,n)=1 \,\}$\\
$M_l^{\text{ST}}(b,n) \gets \mathbb{I}\!\big(t^\star(b,n)\le \phi_l\big)\ \in\ \mathbb{B}^{B\times N_{l-1}}$

\textbf{(4) Fallback to keep at least one token (per-sample)}\\
\For{$b=1$ \KwTo $B$}{
  \If{$\sum_{n} M_l^{\text{ST}}(b,n)=0$}{
    $n_{\max}\gets \arg\max_{n}\sum_{t} \bar{A}_l(b,t,n)$\\
    $M_l^{\text{ST}}(b,n_{\max})\gets 1$
  }
}

\textbf{(5) Dense reindexing \& padding (batched)}\\
$\mathcal{S}_l^{(b)} \gets \{\, n\mid M_l^{\text{ST}}(b,n)=1 \,\}$,\quad $N_l^{(b)}\gets|\mathcal{S}_l^{(b)}|$\\
$\tilde{O}_l[b,t,c,i] \gets O_l[b,t,c,\mathcal{S}_l^{(b)}[i]]\quad (i=1,\dots,N_l^{(b)})$\\
$N_l \gets \max_b N_l^{(b)}$\\
$P_l \gets \operatorname{Pad}(\tilde{O}_l, N_l)\ \in\ \mathbb{B}^{B\times T\times C\times N_l}$\\
$\mathrm{Mask}_l[b,n]\gets \mathbb{I}(n\le N_l^{(b)})\ \in\ \mathbb{B}^{B\times N_l}$

\Return $P_l,\ \mathrm{Mask}_l$
\end{algorithm}

\subsection{Spike-Guided Spatio-Temporal Token Pruner}\label{sec:DST-TP}
To leverage the inherent sparsity of SNNs while reducing the computational complexity of the SSM, which scales linearly with respect to the sequence length, we design a \textsc{Spike-Guided Spatio-Temporal Token Pruner} (SST-TP). 
The pruner operates after the linear projection layer and before entering the core SmolMamba block where the SSM scan is performed. 
Our strategy jointly considers both the spatial spike activity and the temporal salience of tokens. 

For the spike input 
$O_{l} \in \mathbb{B}^{B \times T \times C \times N_{l-1}}$,
where $B$, $T$, $C$, and $N_{l-1}$ denote the batch size, timestep, channel, and token dimensions, respectively,
we first compute the raw channel-wise spike activity:
\begin{equation}
    \bar{A}_{l}
    = \frac{1}{C}\sum_{c=1}^{C} O_{l}[:,:,c,:]
    \;\in\;
    \mathbb{R}^{B \times T\times N_{l-1}}.
\label{eq:spike_activity}
\end{equation}

To obtain a relative salience measure across tokens, 
we perform Z-score normalization along the token dimension for each $(b,t)$:
\begin{equation}
\label{eq:zscore}
\tilde{A}_{l}[b,t,n]
=
\frac{
{A}_{l}[b,t,n]
-
\mu_{l}[b,t]
}{
\sigma_{l}[b,t] + \varepsilon
},
\end{equation}
where 
$\mu_{l}[b,t]$ and $\sigma_{l}[b,t]$ denote the mean and standard deviation of 
${A}_{l}[b,t,:]$ across tokens,
and $\varepsilon$ is a small constant to ensure numerical stability.
After normalization, $\tilde{A}_l$ has zero mean, which enables the use of a fixed activity threshold.

A token is regarded as spatially significant if its Z-score exceeds zero.
The spatial mask is therefore defined as
\begin{equation}
\label{eq:spatial_mask_z}
M_{l}^{\mathrm{spatial}}
=
\mathbb{I}\!\left(
\tilde{A}_{l} > 0
\right)
\in
\mathbb{B}^{B \times T \times N_{l-1}},
\end{equation}
where $\mathbb{I}(\cdot)$ is the indicator function applied element-wise.

In addition, neuroscience studies indicate that neural coding in sensory cortices depends on the first-spike latency: stimuli with higher salience elicit shorter times to first spike, allowing faster responses~\cite{van2001rate, stanojevic2024high}. Motivated by this principle, we define a spatiotemporal Boolean mask ${M_{l}^{\text{ST}}}\in \mathbb{R}^{B \times N_{l-1}}$ for final token pruning, which retains tokens exhibiting at least one significant spike before a specified time threshold $\phi_l \in [1, T]$, formulated as:
\begin{equation} {M_{l}^{\text{ST}}}=\mathbb{I}(\arg \min _{t}\left\{t \mid {M_{l}^{\text{spatial}}}\left[:,t,:\right]=1\right\}\leq\phi_{l}). \label{ed:temporal_mask}
\end{equation}

The overall pruning behavior is visualized in Fig.~\ref{fig:dsttp}. Formally, the pruning operation at layer $l$ is defined as a dense reindexing along the token dimension:
\begin{equation}
\begin{aligned}
&\mathcal{S}_l[b] = \big\{\, n \in \{1,\dots,N_{l-1}\} \;\big|\; M_{l}^{\text{ST}}[b,n] = 1 \,\big\}, \\
&N_{l}[b] = |\mathcal{S}_{l}[b]|, \\
&\tilde{O}_l[b,t,c,i] = O_{l}\!\left[b,t,c,\mathcal{S}_l[b][i]\right],
\end{aligned}
\label{eq:prune_dense}
\end{equation}
where $\mathcal{S}_l[b]$ denotes the ordered index set of retained tokens for the $b$-th sample, 
$N_l[b]$ is the corresponding number of preserved tokens, and $i = 1,\ldots, N_l[b]$. 
This dense reindexing operation explicitly discards pruned tokens 
and compacts the remaining ones into a contiguous sequence $\tilde{O}_l[b] \in \mathbb{B}^{T \times C \times N_l[b]}$, 
ensuring that token order is preserved while inactive tokens are entirely removed.

Since each sample may yield a different token length after pruning, 
we align all sequences within a batch by zero-padding them to the maximum retained length:
\begin{equation}
\begin{aligned}
&N_l = \max_{b} N_l[b], \\
&P_l = \operatorname{Pad}\!\left(\tilde{O}_l, N_l\right)\in \mathbb{B}^{B \times T \times C \times N_{l}}, \\
&\text{Mask}_l[b,n] = \mathbb{I}\!\left(n \le N_l[b]\right),
\quad \text{Mask}_l \in \mathbb{B}^{B\times N_l}.
\end{aligned}
\label{eq:padding_mask}
\end{equation}

The resulting $P_l$ is then fed into the SSM module. 
The binary padding mask $\text{Mask}_l$ ensures that padded positions do not contribute to any further computation, maintaining numerical stability and preserving the intended sparsity.

\paragraph{Training behavior}
The spatial threshold (zero Z-score) and latency threshold $\phi_l$ are deterministic hyperparameters shared between training and inference. 
SST-TP is activated in both phases, so the SSM is always optimized under the same dynamically pruned token distributions, and gradients are propagated only through the surviving tokens.

\subsection{SmolMamba Block}
\label{sec:smolmamba}
Mamba~\cite{gu2024mamba} efficiently models long-range dependencies using structured state-space models (SSMs). However, these models evolve hidden states continuously in real-valued space, whereas SNNs propagate discrete membrane potentials through spike events. As a result, the two paradigms differ fundamentally in their temporal representation and update granularity.

To reconcile this discrepancy, we introduce the \textsc{SmolMamba Block},
which integrates spiking neurons into the Mamba architecture to enable
event-driven state-space modeling. However, the resulting spike representations still contain redundant spatio-temporal tokens. To address this, we further introduce a progressive \emph{molting} process—analogous to biological molting—where redundant
tokens are gradually removed across layers.

Following the bidirectional modeling design of Vision Mamba~\cite{zhuVisionMambaEfficient2024}, each SmolMamba block performs bidirectional state scans through two SSM paths. Given the input feature $X_{l-1}\in\mathbb{R}^{B\times T \times C\times N_{l-1}}$, we first apply a convolution–batch normalization–spiking activation sequence to obtain spike-driven representations:
\begin{equation}
O_l =
\mathcal{SN}\!\left(\mathrm{BN}\!\left(\operatorname{Conv2d}(X_{l-1})\right)\!\right),\quad
l = 1,2,\dots,L.
\end{equation}

Next, a pruning step is performed via the proposed Spike-Guided Spatio-Temporal Token Pruner (SST-TP, Sec.~\ref{sec:DST-TP}), which adaptively filters spike activations and removes redundant tokens:
\begin{equation}
(P_l, \operatorname{Mask}_l) =
\operatorname{SST\!-\!TP}_l(O_l).
\end{equation}

Here $P_l\in\mathbb{R}^{B\times T\times C\times N_l}$ denotes the pruned token sequence with $N_l \le N_{l-1}$.

The pruned spike-encoded sequence is then projected through a depthwise convolution to enhance local temporal interactions:
\begin{equation}
(H_l^{f}, H_l^{b}) =
\mathcal{SN}\!\left(
\operatorname{DWConv}(P_l,\operatorname{Reverser}(P_l))
\right).
\end{equation}

Subsequently, two structured SSM modules perform forward and backward state scans for bidirectional context modeling:
\begin{equation}
\label{eq:SSM}
(Z_l^{f}, Z_l^{b}) =
\mathcal{SN}\!\left(
\operatorname{SSM}(H_l^{f}, H_l^{b})
\right).
\end{equation}

The outputs of the two branches are fused and modulated using the pruning mask $\operatorname{Mask}_l$:
\begin{equation}
F_l =
(Z_l^{f} + \operatorname{Reverser}(Z_l^{b}))
\odot
\operatorname{Mask}_l .
\end{equation}

The fused representation is then re-normalized and passed through a spiking neuron layer, followed by addition with the masked input as a residual connection:
\begin{equation}
\begin{aligned}
U_l =
&\mathcal{SN}\!\left(
\mathrm{BN}\!\left(\operatorname{Conv2d}(F_l)\right)
\right) \\
&+ X_{l-1} \odot \operatorname{Mask}_l .
\end{aligned}
\end{equation}

Finally, a spiking MLP head (\textsc{SMLP})—comprising linear projection, batch normalization, and spiking activation—is applied to refine token-level representations and produce the final output:
\begin{equation}
X_l =
\operatorname{SMLP}(U_l) + U_l .
\end{equation}

Unlike the vanilla Mamba \cite{gu2024mamba}, we remove the gating branch to eliminate redundant dense activations. The binary nature of spike generation naturally provides an implicit gating mechanism, while the SST-TP module serves as an explicit pruning mechanism that progressively removes inactive tokens across layers. This design forms a differentiable bridge: the continuous SSM backbone captures long-range temporal dependencies, while the discrete spiking dynamics enforce sparse event-driven computation. The hierarchical pruning process further produces temporally aware and energy-efficient visual representations.

\subsection{Mask-aware Global Average Pooling}
\label{sec:maskaware}

Dynamic token pruning results in variable-length token sequences across samples,
which invalidates the assumptions behind conventional global pooling.
To obtain unbiased global representations under varying token counts,
we adopt a \textsc{Mask-aware Global Average Pooling (MGAP)} operator that aggregates features
only over valid (unmasked) tokens.

Given the final-layer feature map 
$X_L\!\in\!\mathbb{R}^{B\times T\times C\times N_L}$ 
and the corresponding binary mask 
$M_L\!\in\!\mathbb{B}^{B\times1\times1\times N_L}$,
the pooled representation for each sample $b$ and channel $c$ is computed as:
\begin{equation}
P_{b,c}=
\frac{
\sum_{t,n} M_{b,1,1,n}\, X_{b,t,c,n}
}{
\sum_{n} M_{b,1,1,n}
},
\quad P\in\mathbb{R}^{B\times C}.
\end{equation}

This operation ensures that only active tokens contribute to the global representation,
while pruned or padded positions are entirely excluded.
The resulting pooled vector $P$ is then passed to the linear 
\textsc{Classification Head (CH)} for prediction.
The mask-aware pooling stabilizes training and prevents representation bias 
arising from dynamically varying token lengths within the SmolMamba backbone.

\section{Experiment}

\begin{table*}[!htbp]
\centering
\caption{Hyper-parameters for image classification on ImageNet-1K and CIFAR10/100.}
\label{table_train_imagenet_detail}
\begin{tabular}{l|c|c|c|c}
\hline
Hyper-parameter     & ImageNet  & CIFAR10/100  &CIFAR10DVS &DVS128 \\ \hline
Timestep         & 4         & 4       &16   &16            \\
Epochs              & 300      & 300     & 106    & 200            \\
Resolution          & 224$\times$224   &32$\times$32 &128$\times$128    &128$\times$128       \\
Batch size          & 128          &128      &16   &16    \\
Optimizer           & AdamW           & AdamW   & AdamW & AdamW       \\
Base Learning rate  & 2.5e-4         & 1e-3   & 1e-3  & 1e-3  \\
Learning rate decay & Cosine     & Cosine & Cosine & Cosine\\
Warmup eopchs       & 30      & 30     &10         &10       \\
Weight decay        & 0.05    & 0.05   &0.06   &0.06 \\
Mixup               & 0.8     & 0.8  & 0.5  & 0.5\\
Cutmix              & 1.0    & 1.0   & 0.0  & 0.0  \\
Mixup-off epoch               & None     & 200  & None & None \\
Label smoothing     & 0.1     & 0.1  & 0.1   & 0.1   \\ \hline
\end{tabular}
\end{table*}

\begin{table*}[t]
\centering
\caption{Results on ImageNet-1K.}
\begin{tabular}{lccccc} 
\toprule
Methods & Architecture & Param (M) & Time Step & Power (mJ) & Acc. (\%) \\ 
\midrule
Hybrid training\cite{rathi2020enablingdeepspikingneural} & ResNet-34 & 21.8 & 250 & - & 61.5 \\
TET\cite{deng2022temporalefficienttrainingspiking} & SEW-ResNet-34 & 21.8 & 4 & - & 68.0\\
Spiking ResNet\cite{hu2021spiking}& ResNet-50 & 25.6 & 350 & 70.9 & 72.8 \\
STBP-tdBN\cite{zheng2021going} & Spiking-ResNet-34 & 21.8 & 6 & 6.4 & 63.7 \\
\cmidrule{2-6}
\multirow{2}{*}{SEW-ResNet\cite{fang2021deep}} & SEW-ResNet-34 & 21.8 & 4 & 4.0 & 67.0 \\
 & SEW-ResNet-152 & 60.2 & 4 & 12.9 & 69.3 \\
\cmidrule{2-6}
\multirow{2}{*}{MS-ResNet\cite{hu2024advancing}} & MS-ResNet-34 & 21.8 & 4 & 5.1 & 69.4 \\
 & MS-ResNet-104 & 77.3 & 4 & 10.2 & 75.3 \\
  \midrule
\multirow{2}{*}{Spikformer\cite{zhou2023spikformer}} & Spikformer-8-384 & 16.8 & 4 & 7.7 & 70.2 \\
 & Spikformer-8-512 & 29.7 & 4 & 11.6 & 73.4 \\
\cmidrule{2-6}
\multirow{2}{*}{Spikingformer\cite{zhou2023spikingformer}} & Spikingformer-8-384 & 16.8 & 4 & 4.7 & 72.5 \\
 & Spikingformer-8-512 & 29.7 & 4 & 7.5 & 74.8 \\
\cmidrule{2-6}
\multirow{3}{*}{Spike-driven Transformer\cite{yao2023spike}} 
& Spike-driven Transformer-8-384 & 16.8 & 4 & 3.9 & 72.3 \\
  & Spike-driven Transformer-8-512 & 29.7 & 4 & 4.5 & 74.6 \\
 & Spike-driven Transformer-10-512 & 36.0 & 4 & 5.5 & 74.7 \\

\cmidrule{2-6}
\multirow{3}{*}{SNN-ViT\cite{wangspiking}} 
& SNN-ViT-8-256 & 13.7 & 4 & 14.28 & 74.66 \\
  & SNN-ViT-8-384 & 30.4 & 4 & 20.83 & 76.87 \\
 & SNN-ViT-8-512 & 53.7 & 4 & 35.75 & 80.23 \\
 
  \midrule
\multirow{2}{*}{Vision Mamba\cite{zhou2023spikingformer}} & Vim-T & 7 & 1 & 6.7 & 76.1 \\
 & ViM-S & 26 & 1 & 23.4 & 80.3 \\
 \multirow{2}{*}{Dynamic Vision Mamba\cite{zhou2023spikingformer}} & Vim-T + DyVM & 7 & 1 & 5.8 & 75.2 \\
 & ViM-S + DyVM & 27 & 1 & 15.1 & 78.8 \\
 \midrule
\multirow{4}{*}{Vision SmolMamba (Ours)} 
 &  Vision SmolMamba-8-384 & 16.1 & 4 & 1.6 & \textbf{73.5} \\
 &  Vision SmolMamba-6-512 & 22.4 & 4 & 1.7 & \textbf{75.3} \\
 &  Vision SmolMamba-8-512 & 28.4 & 4 & 2.0 & \textbf{76.8} \\
 &  Vision SmolMamba-10-512 & 34.4 & 4 & 2.6 & \textbf{78.9} \\
\bottomrule
\end{tabular}
\label{tab:imagenet_comparison}
\end{table*}


We conducted experiments on both static datasets, including CIFAR10/100 \cite{krizhevsky2009cifar10} and ImageNet~\cite{93443fe0aca44aa4b4f773517d12ee30}, as well as event-based datasets such as CIFAR10-DVS \cite{li_cifar10-dvs_2017} and DVS128 Gesture \cite{amir2017low}. Unless otherwise specified, all experiments are conducted on an AMD EPYC 7282 16-core CPU and distributed across 8 NVIDIA A100 GPUs. Details of the training hyper-parameters are provided in Tab.~\ref{table_train_imagenet_detail}.

Following prior neuromorphic computing studies~\cite{6757323,zhou2023spikformer,yao2023spike,zhou2023spikingformer}, we report \emph{estimated energy cost} rather than hardware-measured power. Our estimation focuses on the intrinsic compute energy and does not include memory access or communication overhead. The energy values are derived from the number of elementary operations executed during inference together with the corresponding energy cost per operation. This estimation protocol is widely adopted in prior SNN energy analyses and enables a fair relative comparison of computational efficiency across architectures. The detailed estimation procedure is described in Sec.~\ref{sec:energy}.

\subsection{ImageNet}
Tab.~\ref{tab:imagenet_comparison} reports a comprehensive comparison on ImageNet-1K. 
Across all model scales, Vision SmolMamba exhibits a consistently superior  
accuracy–energy trade-off.

\paragraph{Compared to Transformer-based SNNs}
Under the same architectural setting (8--512), Vision SmolMamba reduces the 
\emph{estimated energy cost} from 7.7--11.6 mJ (Spikformer) and 7.5 mJ (Spikingformer) 
down to only 2.0 mJ, corresponding to a substantial reduction in energy cost.
Even compared with the highly optimized Spike-driven Transformer 
(4.5 mJ for 8--512), Vision SmolMamba still achieves a 
2.25$\times$ reduction in estimated energy cost while improving accuracy.

\paragraph{Compared to ANN Mamba variants}
Although ViM-S and ViM-S+DyVM achieve stronger raw accuracy, their estimated energy usage 
(23.4 mJ and 15.1 mJ) is substantially higher. Vision SmolMamba-8-512 consumes only 
8.5\%--13.2\% of their energy cost, equivalent to a 7.6$\times$--11.7$\times$ improvement 
in energy efficiency, while trailing in accuracy by merely 3.5\% and 2.0\%.

\paragraph{Scaling benefits}
Energy savings persist as the model scales. 
Vision SmolMamba-10-512 reaches an accuracy of 78.9\% with only 2.6 mJ 
estimated energy cost, representing a 2.1$\times$ improvement in accuracy-per-mJ 
and a 4.2\% accuracy gain over Spike-driven Transformer-10-512 
despite using fewer parameters.

\begin{figure}[htbp]
\centering
\includegraphics[width=1\linewidth]{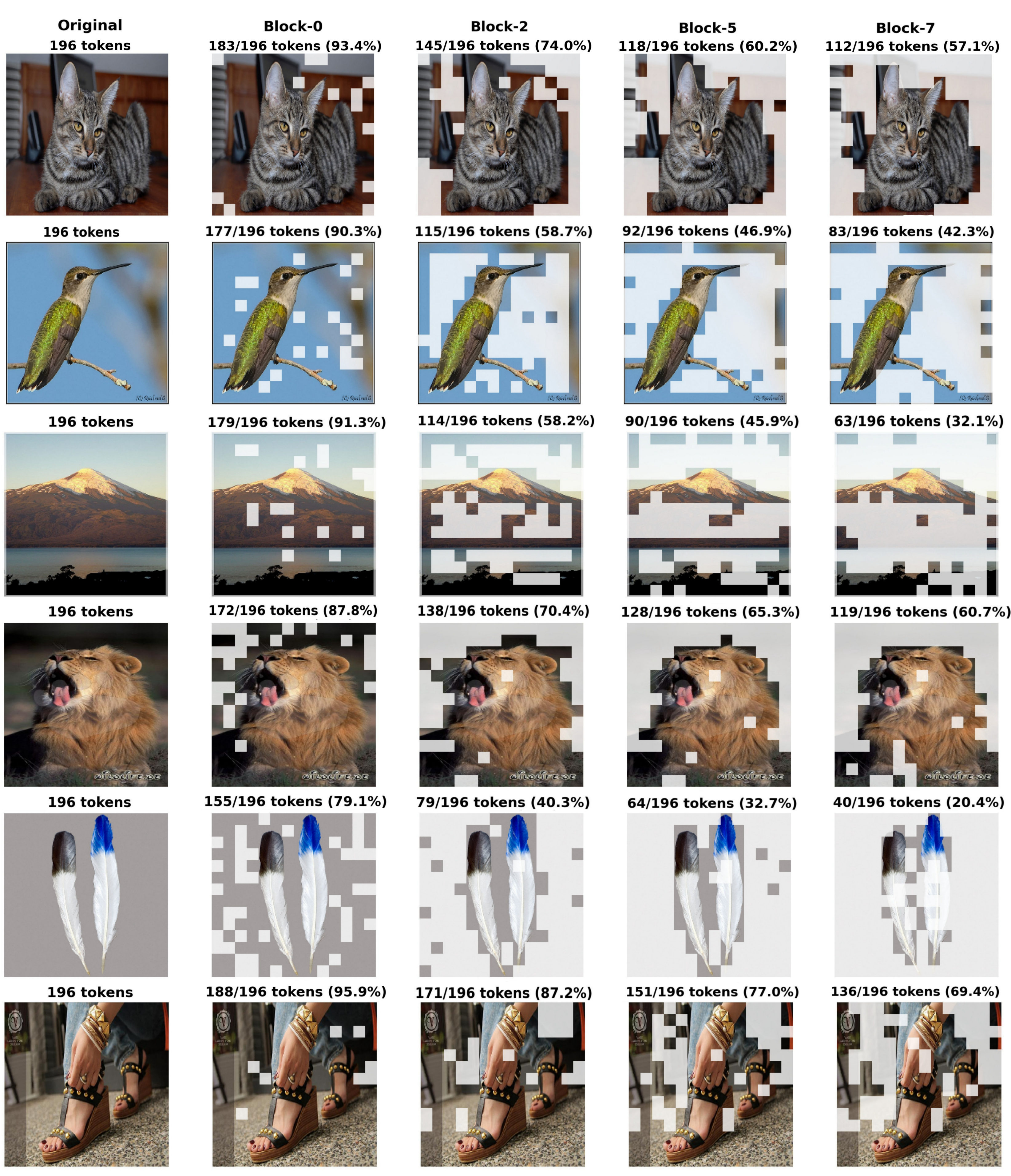}
\caption{Visualization of token pruning results in ImageNet-1k. In each group of images, we show the original image, along with its retained tokens of each pruning stage.  Pruned tokens are mostly from low-activity areas, implying their redundancy.}
\label{fig:token_pruning_1}
\end{figure}

\paragraph{Token pruning visualization}
As shown in Fig.~\ref{fig:token_pruning_1}, the retained tokens at each stage show that spatially inactive regions are progressively suppressed, while tokens with strong spike activity are consistently preserved. 
Across different images, the model exhibits distinct routing patterns, indicating that Vision SmolMamba dynamically adapts its computational path to the input. 
These visualizations demonstrate that the model effectively removes redundant tokens while preserving salient representations essential for long-range spiking state-space modeling.

Overall, Vision SmolMamba achieves at least $2\times$ lower estimated energy cost than the compared baselines on ImageNet-1K while maintaining comparable or better accuracy, demonstrating the effectiveness of its spiking state-space formulation and spike-guided spatio-temporal pruning strategy.

\begin{table*}[htbp]
\centering
\caption{Experimental Results on CIFAR10/100, DVS128 Gesture and CIFAR10-DVS.}
\label{tab:exp_results}
\begin{tabular}{l cc cc cc cc}
\hline
\multirow{2}{*}{Methods} & \multicolumn{2}{c}{CIFAR10-DVS} & \multicolumn{2}{c}{DVS128 Gesture} & \multicolumn{2}{c}{CIFAR-10} & \multicolumn{2}{c}{CIFAR-100} \\ \cline{2-9} 
 & T & Acc. (\%) & T & Acc. (\%)& T & Acc. (\%) & T & Acc. (\%) \\
\hline
tdBN \cite{zheng2021going} & 10 & 67.8 & 40 & 96.9 & 6 & 93.2 & - & - \\
PLIF \cite{fang2021incorporating} & 20 & 74.8 & 20 & 97.6 & 8 & 93.5 & - & - \\
Dspike \cite{li2021differentiable} & 10 & 75.4 & - & - & 6 & 94.3 & 6 & 74.2 \\
DSR \cite{mengTrainingHighperformanceLowlatency2022}& 10 & 77.3 & - & - & 20 & 95.4 & 20 & 78.5 \\
Spikformer\cite{zhou2023spikformer} & 16 & 80.9 & 16 & 98.3 & 4 & 95.5 & 4 & 78.2 \\
SDT\cite{yao2023spike} & 16 & 80.0 & 16 & 99.3 & 4 & 95.6 & 4 & 78.4 \\
SAC\cite{liu2023spike} & - & - & 10 & 95.1 & 6 & 93.7 & - & - \\
SparseSpikformer\cite{liu2024sparsespikformer} & 16 & 79.1 & 16 & 98.3 & 4 & 95.2 & 4 & 77.8 \\
Ternary Spike\cite{guo2024ternary} & 10 & 79.8 & - & - & 4 & 95.0 & 4 & 74.0 \\
TCJA-SNN\cite{zhu2024tcja} & 10 & 80.7 & 20 & 99.0 & 4 & 95.6 & 4 & 77.7 \\
FSTA-SNN\cite{yu2025fsta} & 16 & 82.7 & - & - & 4 & 94.7 & 4 & 73.4 \\
P-SpikeSSM\cite{bal2024p} & - & - & - & - & 64 & 78.3 & - & - \\
SpikingSSM\cite{shen2025spikingssms} & - & - & - & - & 64 & 84.5 & - & - \\
\hline
Vision SmolMamba (Ours) & 16 & 82.2 & 16 & 100 & 4 & 95.7 & 4 & 79.0 \\
\hline
\end{tabular}
\end{table*}

\begin{table}[htbp]
\centering
\small
\setlength{\tabcolsep}{4pt}
\caption{Accuracy and inference throughput comparison on CIFAR datasets. 
Throughput is measured as images processed per second on a single NVIDIA V100 GPU.}
\label{throughput}
\begin{tabular}{lcccc}
\hline
\multirow{2}{*}{Model} & \multicolumn{2}{c}{CIFAR10} & \multicolumn{2}{c}{CIFAR100} \\ 
\cmidrule(lr){2-3}\cmidrule(lr){4-5}
& Acc. (\%) & Img/s & Acc. (\%) & Img/s \\
\hline
Spikformer \cite{zhou2023spikformer} & 95.5 & 950 & 78.2 & 1049 \\
SparseSpikformer \cite{liu2024sparsespikformer} & 95.2 & 1293 & 77.8 & 1156\\
Vision SmolMamba (Ours)   & 95.7 & 1602 & 79.0 & 1651\\
\hline
\end{tabular}
\end{table}

\begin{table}[t]
\centering
\caption{Ablation study of Vision SmolMamba-2-256 on CIFAR-10.}
\label{tab:ablation_detailed}
\begin{tabular}{lcc}
\toprule
Modules                          &  Power ($\mu$J) & Acc. (\%) \\
\midrule
SSA \cite{zhou2023spikformer} (Spikformer)      & 91.4       & 93.7 \\
Spiking Mamba                   & 48.2        & 94.7 \\
Spatial-Only TP                & 36.4        & 94.4 \\
Temporal-Only TP               & 45.8        & 94.3 \\
\midrule
SST-TP (w/o  Z-score normaliza)   & 47.0        & 94.6 \\
Full SST-TP                    &  31.9        & 94.6 \\
\bottomrule
\end{tabular}
\end{table}

\subsection{CIFAR and DVS}
\paragraph{Accuracy Results}
As shown in Tab.~\ref{tab:exp_results}, we compared our method with the prior SoTA works on four datasets. CIFAR10-DVS and DVS128 Gesture are both neuromorphic datasets, where CIFAR10-DVS is converted from the static CIFAR dataset by shifting image samples to be captured by the DVS camera, and DVS128 Gesture is an event-based gesture recognition dataset. We follow the data augmentation applied in \cite{li2022neuromorphic}. On Vision SmolMamba-2-256 (2 blocks and 256 channels), we achieve 82.2\% on CIFAR10-DVS and 100\% on DVS128 Gesture using 106 and 200 epochs respectively. Comparable results to SOTA on static datasets like CIFAR10 (95.8\%) and CIFAR100 (79.0\%) are reached on our Vision SmolMamba-2-512.

To provide a comprehensive evaluation against recent state space models adapted for SNNs, we reproduce SpikingSSM and pSpikeSSM as baseline comparisons. Originally, these models were designed for the Long Range Arena (LRA) benchmark. For a fair and direct comparison on standard 2D visual tasks, we align their input paradigm with modern vision architectures by equipping them with a $4 \times 4$ 2D patch embedding (preserving 3-channel RGB information). Consequently, a $32 \times 32$ image is divided into $8 \times 8 = 64$ sequence tokens. It is crucial to clarify the definition of the time step $T$ in Tab.~\ref{tab:exp_results}. For traditional spatial-based SNNs (including our Vision SmolMamba), $T$ denotes the extra simulation time steps applied to the static spatial tokens (e.g., $T=4$). Conversely, for sequence-based SNNs like SpikingSSM and pSpikeSSM, temporal dynamics natively unfold along the sequence dimension without an extra simulation loop. Therefore, their equivalent time step $T$ is defined by their sequence length, which is 64 in this patch-based setting. As demonstrated in Tab.~\ref{tab:exp_results}, these 1D sequence-oriented baselines fail to achieve high accuracy on CIFAR10. This clearly highlights that our architecture possesses a superior spatial representation capacity for visual tasks.

\paragraph{Throughput Comparison}
To further evaluate the computational efficiency of our architecture, we report the inference throughput on CIFAR10 and CIFAR100 in Tab.~\ref{throughput}. Throughput is measured as the number of images processed per second on a single NVIDIA V100 GPU under the same batch size and inference settings. Compared with Spikformer and SparseSpikformer, Vision SmolMamba achieves consistently higher throughput while maintaining competitive or superior accuracy. In particular, Vision SmolMamba improves throughput from 950.28 to 1602.23 on CIFAR10 and from 1049.55 to 1651.42 on CIFAR100.

This efficiency gain primarily stems from the architectural differences among these models. Spikformer adopts a spiking transformer backbone with spike-form attention, while SparseSpikformer introduces firing-rate-based token pruning to reduce redundant token interactions. In contrast, Vision SmolMamba replaces the attention mechanism with a spiking state-space backbone and further incorporates spatio-temporal token pruning. This design enables linear-time token processing and avoids the quadratic token interactions required by attention-based spiking transformers, resulting in significantly improved inference throughput.

\subsection{Ablation Study}

This section provides extended ablation results of Vision SmolMamba. Unless otherwise stated, experiments are conducted on CIFAR-10 using Vision SmolMamba-2-256 and on ImageNet-1K using Vision SmolMamba-8-512.

To evaluate the contribution of each design component, we perform ablation studies on the CIFAR-10 dataset by removing or modifying key modules from the full Vision SmolMamba architecture. All variants share identical hyperparameter settings to ensure a fair comparison, except for the specific architectural differences described below.

The evaluated variants are summarized as follows:

\begin{itemize}
\item \textbf{SSA (Spikformer):} Replacing the SmolMamba block with the spiking self-attention (SSA) mechanism used in Spikformer \cite{zhou2023spikformer}, without applying any token pruning.

\item \textbf{Spiking Mamba (w/o Purning):} Removing the proposed SST-TP module from Vision SmolMamba while preserving the spiking state-space backbone, resulting in a variant without token pruning.

\item \textbf{Spatial-Only TP:} Applying token pruning based solely on spatial spike activation strength within the Vision SmolMamba architecture.

\item \textbf{Temporal-Only TP:} Applying token pruning based solely on temporal first-spike latency within the Vision SmolMamba architecture.

\item \textbf{SST-TP (w/o Z-score normalization):} Applying the SST-TP mechanism within Vision SmolMamba but without the Z-score normalization used to balance spatial and temporal scores.

\item \textbf{Full SST-TP:} The complete Vision SmolMamba model using the proposed spiking SSM backbone together with spike-guided spatio-temporal token pruning (SST-TP) with Z-score normalization.
\end{itemize}

\paragraph{Effect of SST-TP}

To verify the effectiveness of the proposed SST-TP mechanism, we conduct ablation experiments on the CIFAR-10 dataset. The SSA module from Spikformer~\cite{zhou2023spikformer} is adopted as an attention-based baseline. The results are reported in Tab.~\ref{tab:ablation_detailed}, where Power denotes the estimated energy cost of components other than SPS.

When SST-TP and its variants are applied to the Spiking SSM backbone, all configurations achieve higher accuracy than the SSA baseline while maintaining lower energy cost. Compared with the no-pruning Spiking Mamba, the full SST-TP model introduces only a marginal accuracy decrease of 0.1\% while reducing the computational cost by approximately 1.5$\times$.
This reduction originates from the selective token pruning performed before the spiking SSM computation. By removing redundant tokens, SST-TP effectively reduces the number of spike operations, resulting in lower SOP counts and consequently reduced energy cost.
The spatial-only variant retains tokens solely according to spatial spike intensity, ignoring temporal spike latency, which leads to inferior accuracy compared with SST-TP. The temporal-only variant retains tokens based on early spike occurrence regardless of spike magnitude. This strategy also underperforms SST-TP and fails to achieve substantial computational reduction because SNNs naturally generate numerous low-intensity spike events.

Additionally, we evaluate SST-TP without Z-score normalization in Eq.~\eqref{eq:zscore}. In this case, tokens compete using raw activity values. As training progresses, the pruning mechanism degenerates, causing most tokens to become either fully pruned or fully retained. In contrast, the normalized criterion maintains stable pruning ratios and consistently yields better accuracy–efficiency trade-offs.

\begin{figure*}[!t]
  \centering
  \subfloat[]{
    \includegraphics[width=1\linewidth]{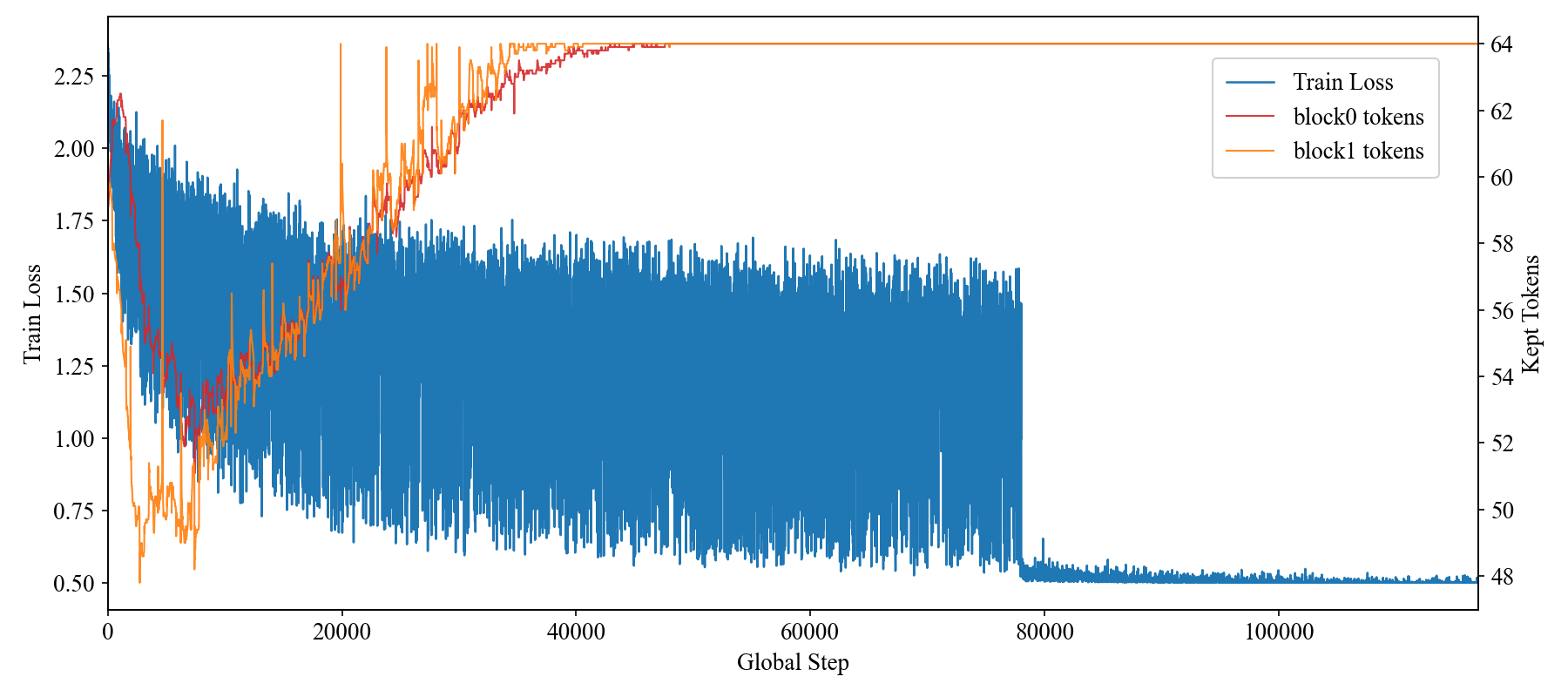}
    \label{fig:cifar10loss1}
  }
  \hfill
  \subfloat[]{
    \includegraphics[width=1\linewidth]{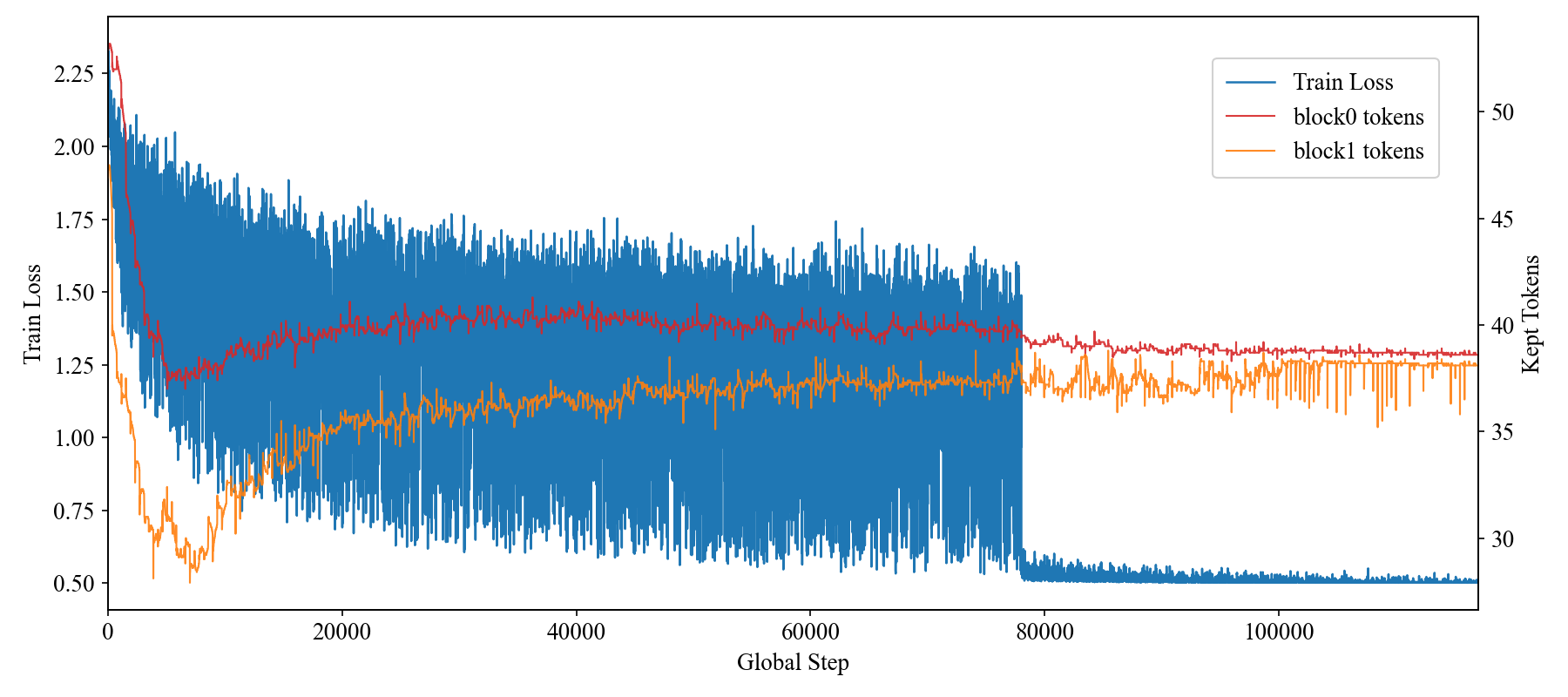}
    \label{fig:cifar10loss2}
  }
  \caption{Train loss and kept tokens of Vision SmolMamba-2-256 on Cifar-10.
(a) Without Z-score normalization, pruning stays inactive early and collapses suddenly.
(b) With Z-score normalization, pruning evolves smoothly and stably.}
  \label{fig:cifar10loss}
\end{figure*}

\paragraph{Effect of Z-score normalization on pruning dynamics in SST--TP}
The effect of Z-score normalization differs substantially between small-scale (CIFAR-10) and large-scale (ImageNet-1K) training. On CIFAR-10, both normalized and unnormalized variants train stably and reach similar accuracy (Fig.~\ref{fig:cifar10loss}). The difference lies in pruning behavior: without normalization, raw spike magnitudes grow during training, allowing increasingly many tokens—including redundant ones—to surpass the fixed threshold. As a result, the pruner gradually loses selectivity and its kept-token ratio approaches the unpruned level. With Z-score normalization, the activity distribution is standardized at each timestep, making the zero threshold consistently represent “above-average” salience and preserving meaningful pruning across training.

\begin{figure*}[!t]
  \centering
  \subfloat[]{
    \includegraphics[width=1\linewidth]{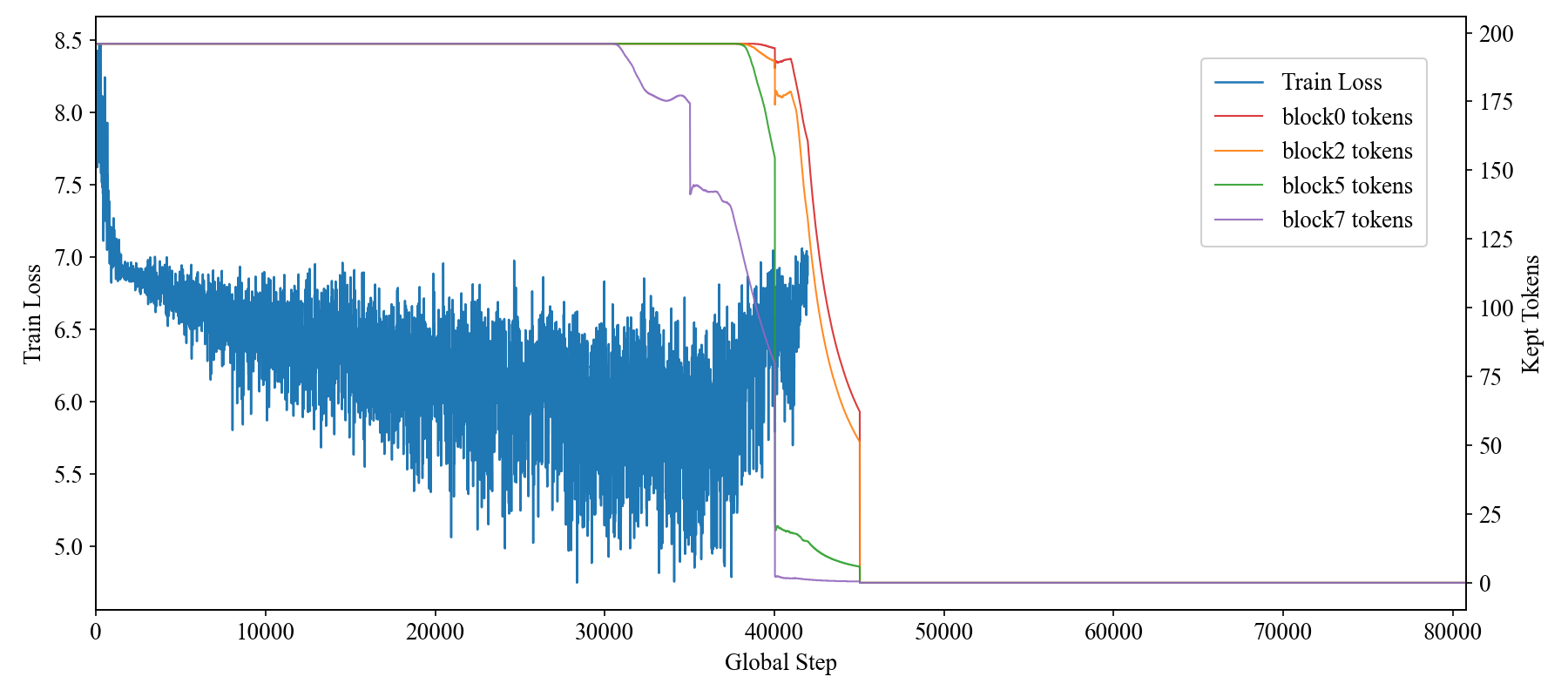}
    \label{fig:imagenet1kloss1}
  }
  \hfill
  \subfloat[]{
    \includegraphics[width=1\linewidth]{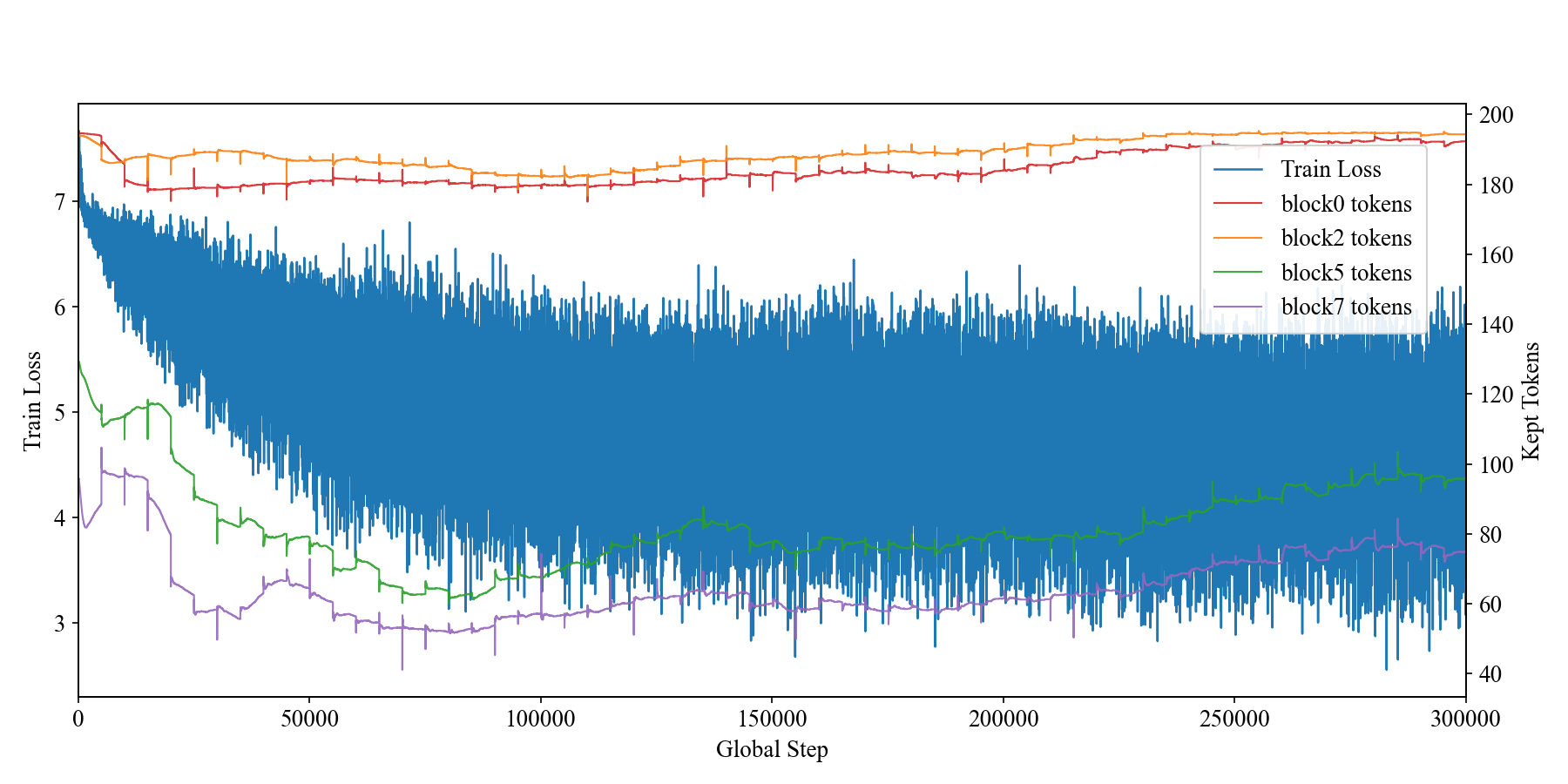}
    \label{fig:imagenet1kloss2}
    }
  \caption{Train loss and kept tokens of Vision SmolMamba-8-512 on ImageNet-1K.
(a) Without Z-score normalization, pruning stays inactive early and collapses suddenly.
(b) With Z-score normalization, pruning evolves smoothly and stably.}
  \label{fig:imagenet1kloss}
\end{figure*}

On ImageNet-1K, the difference becomes significantly more pronounced. The unnormalized variant exhibits unstable pruning dynamics (Fig.~\ref{fig:imagenet1kloss}). Due to the substantially larger activation variance, the fixed threshold alternates between excessive pruning and insufficient pruning, eventually leading to mask collapse and NaN losses.

Tab.~\ref{tab:additional_ablation_detailed} compares three ImageNet-1K configurations to evaluate the stability of Vision SmolMamba on large-scale datasets: (1) a no-pruning baseline (Spiking Mamba), (2) SST-TP without Z-score normalization, and (3) the Z-score–normalized Full SST-TP.

The unnormalized SST-TP variant fails to train. The no-pruning baseline trains stably but retains all tokens and therefore incurs the highest energy cost. In contrast, the normalized Full SST-TP converges reliably, performs selective token pruning, and achieves substantial reductions in token count and SOP with only a minor accuracy decrease.

These results indicate that although Z-score normalization is not strictly required for numerical stability on small datasets, it becomes essential for maintaining pruning selectivity and stable optimization when scaling to large datasets. In this sense, Z-score normalization effectively introduces a scale-invariant pruning criterion that stabilizes spike-driven token pruning.

\paragraph{Effect of Z-score Threshold}
Tab.\ref{tab:st} shows analysis on Z-score thresholds (CIFAR-10, Vision SmolMamba-2-256). When the threshold is 0 it reaches a best trade-off in accuracy and efficiency.

\begin{table}[t]
\centering
\caption{Ablation study of Vision SmolMamba-8-512 on ImageNet-1k.}
\label{tab:additional_ablation_detailed}
\begin{tabular}{lcc}
\toprule
Methods                          &  Power (mJ) & Acc. (\%) \\
\midrule
Spiking Mamba (w/o Purning)        & 3.4       & 78.5 \\
SST-TP (w/o  Z-score normaliza)     & N/V        & 7.5 \\
Full SST-TP                         & 2.0        & 76.8 \\
\bottomrule
\end{tabular}
\end{table}

\begin{table}[htbp]
\centering
\caption{Effect of Z-score Threshold on Tokens and Accuracy.}
\label{tab:st}
\begin{tabular}{cccccc}
\hline
{Z-score threshold} & Block 1  & Block 2  & Acc. (\%) \\
\hline
-1  & 62.1 & 57.5 & 94.6 \\
-0.5  & 51.7 & 48.5 & 95.0 \\
\textbf{0}  & \textbf{38.8} & \textbf{37.8} & \textbf{94.6} \\
0.5  & 29.6 & 13.6 & 93.7 \\
1  & 18.1 & 6.7 & 93.6 \\
\hline
\end{tabular}
\end{table}



\paragraph{Effect of different temporal thresholds}
Tab.~\ref{tab:tt} illustrates the performance under different temporal thresholds $\phi_l$. As $\phi_l$ increases from 1 to 4, the average number of tokens retained in both Block 1 and Block 2 steadily increases. The model has reached an optimal trade-off at $\phi_l=3$, with 94.6\% Acc and about 40\% tokens being pruned.

\begin{table}[htbp]
\centering
\caption{Tokens (Avg.) and Acc of Vision SmolMamba-2-256 with different temporal threshold $\phi_l$ on CIFAR-10.}
\label{tab:tt}
\begin{tabular}{cccccc}
\hline
{$\phi_l$} & Block 1  & Block 2  & Acc. (\%) \\
\hline
1  & 31.3 & 18.4 & 93.7 \\
2  & 37.0 & 30.5 & 94.2 \\
3  & 38.8 & 37.8 & \textbf{94.6} \\
4  & 40.1 & 38.5 & 94.4 \\
\hline
\end{tabular}
\end{table}

\section{Discussion}
\subsection{Complexity of SmolMamba}
\label{sec:complexity}

We provide a detailed per-layer complexity analysis of the proposed \textsc{SmolMamba} architecture, which consists of the Spike-Guided Spatio-Temporal Token Pruner (SST-TP) followed by the spiking SSM backbone.
Let $B$, $T$, $C$, and $N_l$ denote the batch size, timestep, channel, and token dimensions at layer~$l$, respectively, and
$\rho_l=N_l/N_{l-1}\!\in(0,1]$ the keep ratio.

\paragraph{Cost of SST-TP}
Given spike-form input $F_l\!\in\!\mathbb{B}^{B\times T\times C\times N_{l-1}}$,
SST-TP performs three vectorized operations:
(i) channel-wise spike-activity accumulation (Eq.~\ref{eq:spike_activity}), (ii) spatio-temporal mask generation (Eq.~\ref{eq:zscore}--\ref{ed:temporal_mask}), and (iii) dense reindexing (Eq.~\ref{eq:prune_dense}). Counting elementwise operations gives
\begin{equation}
\small
\mathcal{C}_l^{\mathrm{SST\text{-}TP}}
=
\underbrace{\mathcal{O}(B\,T\,C\,N_{l-1})}_{\text{activity}}
+
\underbrace{\mathcal{O}(B\,T\,N_{l-1})}_{\text{masking}}
+
\underbrace{\mathcal{O}(B\,T\,C\,N_l)}_{\text{reindex}}.
\end{equation}
The dominant term arises from activity accumulation,
\begin{equation}
\mathcal{C}_l^{\mathrm{SST\text{-}TP}}\approx\mathcal{O}(B\,T\,C\,N_{l-1}).
\end{equation}
Since all operations are elementwise tensor reductions or boolean scans, they are fully parallelizable on modern GPUs.
Treating $B$, $T$, and $C$ as constants, the arithmetic work scales linearly with respect to the sequence length $N_{l-1}$.
Consequently, its wall-clock cost is negligible compared to the subsequent spiking SSM scan.

\paragraph{Cost of the Spiking SSM (Mamba-style scan)}
After pruning to $N_l$ tokens, the spiking SSM performs bidirectional selective scans \emph{along the token dimension} for each $(b,t,c)$ slice. With a fixed state dimension $S$ (e.g., $S{=}16$), the arithmetic work per layer is
\begin{equation}
\mathcal{C}_l^{\mathrm{SSM}} = \mathcal{O}(2B\,T\,C\,S\,N_l) = \mathcal{O}(N_l).
\end{equation}
Here $B$, $T$, $C$, and $S$ are treated as constants, since they remain fixed across layers and datasets. The parallel depth under the \textsc{s6} scan implementation is $\mathcal{O}(\log N_l)$.

\paragraph{Comparison to attention}
A (Spiking) Transformer layer dominated by self-attention must compute an $N_l{\times}N_l$ affinity matrix:
\begin{equation}
\mathcal{C}_l^{\mathrm{Attn}}=\mathcal{O}(B\,N_l^2\,C),
\qquad
\text{mem}=\mathcal{O}(N_l^2).
\end{equation}
Hence SmolMamba scales linearly with $N_l$, while attention scales quadratically due to the
$N_l{\times}N_l$ affinity computation. The SST-TP molting step further improves efficiency by reducing $N_l$
before the linear-time SSM scan.


\subsection{Estimated Energy Analysis}
\label{sec:energy}

When excluding the energy factors related to hardware fabrication, data access, and memory storage, it is still meaningful to estimate the intrinsic computational energy cost of different network architectures.
Following the CMOS energy model of \cite{6757323}, one 32-bit multiply–accumulate (MAC) operation consumes about
$E_{\mathrm{MAC}}\!=\!4.6$\,pJ, while a single accumulate (AC) operation costs $E_{\mathrm{AC}}\!=\!0.9$\,pJ on a 45\,nm process, satisfying $E_{\mathrm{AC}}\!\ll\!E_{\mathrm{MAC}}$. These coefficients are widely adopted in prior SNN energy analyses \cite{panda_2020_toward,qiu2025efficient,shan2025advancing,t_revsnn}.

\paragraph{Energy formulation}
For a Vision SmolMamba network consisting of $L$ stacked \textsc{SmolMamba Blocks}, the total inference energy is estimated based on the number of elementary operations executed during inference. Following prior neuromorphic energy analysis, we decompose the total energy into multiply–accumulate (MAC) and accumulate (AC) operations.

For spiking layers, the number of spike operations (SOPs) can be approximated as
\begin{align}
\text{SOPs}(l) = fr_l \times T \times \text{MACs}(l),
\label{eq:sop}
\end{align}
where $fr_l$ denotes the average firing rate of the input spike train to the $l$-th block, $T$ is the number of simulation timesteps, and $\text{MACs}(l)$ denotes the corresponding dense operation count.

The total inference energy of Vision SmolMamba can then be estimated as
\begin{align}
E_{\text{SmolMamba}}
&= E_{\text{MAC}} \times \text{MACs}^1_{\text{SPS Conv}}
\nonumber \\
&\quad + E_{\text{AC}} \times
\sum_{n=2}^{N} \text{SOPs}^n_{\text{SNN Conv}}
+ \sum_{l=1}^{L} E^l_{\text{Block}},
\label{eq:energy_total}
\end{align}
where the first SPS convolution layer performs MAC operations to encode static RGB images into spike-form, while subsequent spiking convolution layers mainly involve AC operations.

Within each SmolMamba block, most convolutional and fully connected layers perform AC operations due to the binary spike representation. The selective scan module of the SSM introduces a small number of MAC operations arising from multiplications involving the parameters $\bar{\mathbf{A}}$, $\mathbf{C}$, and the hidden state
$\mathbf{h}$. Consequently, the energy consumption of the $l$-th block can be estimated as
\begin{equation}
\small
\begin{aligned}
E^l_{\text{Block}}
&= E_{\text{MAC}} \times \text{MACs}^l_{\text{SSM}} \\
&\quad + E_{\text{AC}} \times
\left(
\text{SOPs}^l_{\text{SNN Conv}}
+ \text{SOPs}^l_{\text{SNN FC}}
+ \text{SOPs}^l_{\text{SSM}}
\right).
\end{aligned}
\label{eq:energy_block}
\end{equation}

\paragraph{Synaptic operation estimation}
Within each SmolMamba block, the dominant event-driven computations arise from the spiking SSM path and the spiking MLP head. Their synaptic operation counts can be approximated as
\begin{equation}
\text{SOPs}^{l}_{\text{SSM}}\!\propto\!
T\,fr_l\,N_l\,C,\qquad
\text{SOPs}^{l}_{\text{SMLP}}\!\propto\!
T\,fr_l\,N_l\,C^2 .
\end{equation}
where $N_l$ denotes the number of active tokens after SST-TP pruning and $C$ denotes the feature dimension. Because the Spike-Guided Spatio-Temporal Token Pruner directly reduces $N_l$ and removes tokens with low spike activity, the effective synaptic operations are significantly reduced. Consequently, the total energy approximately scales with the cumulative keep ratios $\{\rho_l\}$,
\begin{equation}
E_{\mathrm{total}}\ \propto\
\sum_{l=1}^{L}\rho_l\,fr_l,
\label{eq:energy_scale}
\end{equation}
where $\rho_l=N_l/N_{l-1}$ denotes the token keep ratio at layer $l$.

\paragraph{Comparison with attention-based models}
A Spiking Transformer requires explicit $N_l{\times}N_l$ attention maps whose computational cost scales as $\mathcal{O}(N_l^2C)$. In contrast, SmolMamba’s SSM backbone performs linear-time scans $\mathcal{O}(N_lC)$, while the SST-TP step further reduces the effective sequence length before these scans. As a result, SmolMamba achieves favorable energy scaling with respect to both token length and firing sparsity, combining the linear-time efficiency of SSMs with the event-driven computation advantage of spiking neurons.


\section{Conclusion}
This paper presented SmolMamba, an energy-efficient spiking state-space framework that integrates event-driven spiking computation with linear-time selective recurrence. The proposed Spike-Guided Spatio-Temporal Token Pruner (SST-TP) leverages spike activity and first-spike latency to adaptively remove redundant tokens, while the Spiking Molting Mamba Block incorporates these sparse activations into structured state-space dynamics to progressively refine token representations. In addition, a mask-aware pooling strategy stabilizes training under dynamically varying token lengths.

Extensive experiments on ImageNet and neuromorphic benchmarks demonstrate that SmolMamba achieves improved accuracy–efficiency trade-offs, providing a scalable alternative to quadratic attention in spiking vision models. More broadly, the results suggest that combining spike-driven sparsity with selective state-space modeling offers a promising computational paradigm for energy-efficient long-sequence processing. Although this work focuses on visual recognition, the proposed framework can be naturally extended to other sequential modalities such as audio, event streams, and language, which will be explored in future work.

\bibliographystyle{IEEEtran}
\bibliography{references.bib}

\vfill

\end{document}